\documentclass{ceurart}

\usepackage{tabularx}
\usepackage{microtype}
\usepackage{graphicx}
\usepackage{xspace}
\usepackage{tabularx}
\newcommand{\F}{$\textrm{F}_1$\xspace}
\usepackage{hyperref}
\usepackage{url}
\begin{document}

\copyrightyear{2021}
\copyrightclause{Copyright for this paper by its authors.
  Use permitted under Creative Commons License Attribution 4.0
  International (CC BY 4.0).}

\conference{Forum for Information Retrieval Evaluation, December 13-17, 2021, India}

\title{Multi-Task Learning with Sentiment, Emotion, and Target
  Detection to Recognize Hate Speech and Offensive Language}
\author[1,3]{Flor Miriam Plaza-del-Arco}[email=fmplaza@ujaen.es]
\author[2,3]{Sercan Halat}[email=sercanhalat@mu.edu.tr]
\author[3]{Sebastian Padó}[email=pado@ims.uni-stuttgart.de]
\author[3]{Roman Klinger}[email=klinger@ims.uni-stuttgart.de]

\address[1]{Department of Computer Science, Advanced Studies Center in Information and Communication Technologies (CEATIC), Universidad de Jaén, Campus Las Lagunillas, E-23071, Jaén, Spain}
\address[2]{Department of Turkish Language Teaching, Faculty of Education, Mugla Sitki Kocman University, Kötekli, TR-48000 Mugla, Turkey}
\address[3]{Institut für Maschinelle Sprachverarbeitung, University of Stuttgart, Pfaffenwaldring 5b, 70569 Stuttgart, Germany}

\begin{abstract}
  The recognition of hate speech and offensive language (HOF) is
  commonly formulated as a classification task which asks models to
  decide if a text contains HOF. This task is challenging because of
  the large variety of explicit and implicit ways to verbally attack a
  target person or group. In this paper, we investigate whether HOF
  detection can profit by taking into account the relationships
  between HOF and similar concepts: (a) HOF is related to sentiment
  analysis because hate speech is typically a negative statement and
  expresses a negative opinion; (b) it is related to emotion analysis,
  as expressed hate points to the author experiencing (or pretending
  to experience) anger while the addressees experience (or are
  intended to experience) fear. (c) Finally, one constituting element
  of HOF is the (explicit or implicit) mention of a targeted person or
  group. On this basis, we hypothesize that HOF detection shows
  improvements when being modeled jointly with these concepts, in a
  multi-task learning setup. We base our experiments on existing data
  sets for each of these concepts (sentiment, emotion, target of HOF)
  and evaluate our models as a participant (as team IMS-SINAI) in the
  HASOC FIRE 2021 English Subtask 1A: ``Subtask 1A: Identifying Hate,
  offensive and profane content from the post''. Based on
  model-selection experiments in which we consider multiple available
  resources and submissions to the shared task, we find that the
  combination of the CrowdFlower emotion corpus, the SemEval 2016
  Sentiment Corpus, and the OffensEval 2019 target detection data
  leads to an \F=.7947 in a multi-head multi-task learning model based
  on BERT, in comparison to .7895 of a plain BERT model. On the HASOC
  2019 test data, this result is more substantial with an increase by
  2pp in \F (from 0.78\,\F to 0.8\,\F) and a considerable increase in
  recall. Across both data sets (2019, 2021), the recall is
  particularly increased for the class of HOF (6pp for the 2019 data
  and 3pp for the 2021 data), showing that MTL with emotion,
  sentiment, and target identification is an appropriate approach for
  early warning systems that might be deployed in social media
  platforms.
\end{abstract}


\maketitle

\section{Introduction}
\label{sec:intro}
The widespread adoption of social media platforms has made it possible
for users to express their opinions easily in a manner that is visible
to a huge audience. These platforms provide a large step forward for
freedom of expression. At the same time, social media posts can also
contain harmful content like hate speech and offensive language (HOF),
often eased by the quasi-anonymity on social media platforms
\cite{fortuna2018survey}.  The European Commission’s recommendation
against racism and intolerance defines HOF as ``the advocacy,
promotion or incitement of the denigration, hatred or vilification of
a person or group of persons, as well any harassment, insult, negative
stereotyping, stigmatization or threat of such person or persons and
any justification of all these forms of expression -- that is based on
a non-exhaustive list of personal characteristics or status that
includes `race', color, language, religion or belief, nationality or
national or ethnic origin, as well as descent, age, disability, sex,
gender, gender identity, and sexual orientation'' \cite{ecri2015}.

With the number of social media posts rising sharply, purely manual
detection of HOF does not scale. Therefore, there has been a growing
interest in methods for automatic HOF detection. A straight-forward
approach one might consider is to make use of basic word filters,
which use lexicons that contain entries of words that are frequently
used in hate speech \cite{10.1145/3369869}. This approach, however,
has its limitations, given that HOF depends on discourse, the media,
daily politics, and the identity of the target
\cite{Schmidt2017}. It also disregards the different use of
potentially offending expressions across communities
\cite{sap-etal-2019-risk}.

These factors motivate an interest in more advanced approaches as they
are developed in the field of natural language processing (NLP). Most
recent, well-performing systems make use of machine learning methods
to associate textual expressions in a contextualized manner with the
concept of HOF. Mostly, existing models build on top of end-to-end
learning, in which the model needs to figure out this association
purely from the training data (and a general language representation
which originates from self-supervised pretraining of a language
model).

In this paper, we build on the intuition that HOF is related to other
concepts that might help to direct this learning process. Analyzing
the definition above, HOF is potentially related to \emph{sentiment},
\emph{emotion}, and the \emph{target} of hate speech. First, sentiment
analysis is often defined as the task of classifying an opinion
expression into being positive or negative, given a particular target
\cite{rodriguez2019automatic,liu2010sentiment}. HOF is related as it
typically contains a negative expression or, at least, intend. Second,
emotion analysis is concerned with the categorization of text into a
predefined reference system, for instance basic emotions as they have
been proposed by Paul Ekman \cite{ekman1992argument} (fear, joy,
sadness, surprise, disgust, anger). HOF contains expressions of anger
and might cause fear or other emotions in a target group. Finally, the
target is, by definition, a crucial element of hate speech, whether
mentioned explicitly or not.

The concrete research question we test is whether a HOF detection
system can be improved by exploiting existing resources that are
annotated for emotion, sentiment and HOF target, and carrying out
joint training of a model for HOF and these aspects. In building such
a model, the developer has to decide (a) which of these aspects to
include, (b) which corpora to use for training for each aspect, and
(c) how to combine these aspects. We assume a simple multi-task
learning architecture for (c) and perform model selection on the HASOC
FIRE 2019 development data to address (b). Finally, we address
question (a) through our submissions to HASOC FIRE 2021 Shared Task \cite{hasoc2021mergeoverview} subtask
1A\footnote{\url{https://hasocfire.github.io/hasoc/2021/dataset.html}}\cite{hasoc2021overview} %
which asks systems to carry out a binary distinction between
\emph{``non hate/offensive''} and \emph{``hate/offensive''} English
tweets. We find that a combination of all concepts leads to an
improvement by about 2pp in \F on the HASOC 2019 test data, with a
notable increase in recall by 6pp, and an increase by 0.5pp in \F in the
HASOC 2021 test data (with an increase of 3pp in recall).

\section{Related Work}

As argued above, detecting hate and offensive language on Twitter is a
task closely linked to sentiment, emotion analysis, and
target classification. In this section, we introduce these tasks alongside
previous work and also mention some HOF detection shared tasks that
took place in recent years in the NLP community.

\subsection{Emotion Analysis}
\label{sec:ea}
Emotion analysis from text (EA) consists of mapping textual units to a
predefined set of emotions, for instance basic emotions, as they have
been proposed by \citet{ekman1992argument} (anger, fear, sadness, joy,
disgust, surprise), the dimensional model of
\citet{plutchik2001nature} (adding trust and anticipation), or the
discrete model proposed by \citet{shaver1987emotion}
(anger, fear, joy, love, sadness, surprise). Great efforts have been
conducted in the last years by the NLP community in a variety of
emotion research tasks including  emotion intensity prediction
\cite{strapparava2007semeval,mohammad2017wassa,mohammad-bravo-marquez-2017-emotion},
emotion stimulus or cause detection
\cite{chen-etal-2018-joint,oberlander-klinger-2020-token,Xia2019a,DoanDang2021},
or emotion classification
\cite{mohammad-etal-2018-semeval,oberlander2018analysis}. Studying
patterns of human emotions is essential in various applications such
as the detection of mental disorders, social media mining, dialog
systems, business intelligence, or e-learning. An important
application is the detection of HOF, since it is inextricably linked
to the emotional and psychological state of the speaker
\cite{patrick1901psychology}. Negative emotions such as anger, disgust
and fear can be conveyed in the form of HOF. For example, in the text
``I am sick and tired of this stupid situation'' the author feels angry
and at the same time is using offensive language to express that
emotion. Therefore, the detection of negative emotions can be a clue
to detect this type of behavior on the web.

An important aspect of EA is the creation of annotated corpora to
train machine learning models. The availability of emotion corpora is
highly fragmented, not only because of the different emotion theories,
but also because emotion classification appears to be genre- and
domain-specific \cite{oberlander2018analysis}. We will limit the
discussion of corpora in the following to those we use in this
paper. The Twitter Emotion Corpus (TEC) was annotated with labels
corresponding to Ekman's model of basic emotions (anger, disgust,
fear, joy, sadness, and surprise) and consists of 21,051 tweets. It
was automatically labeled with the use of hashtags that the authors
self-assigned to their posts. The grounded emotions corpus created by
\citet{liu2017grounded} is motivated by the assumption that emotions
are grounded in contextual experiences. It consists of 2,557
instances, labeled by domain experts for the emotions of happiness and
sadness. EmoEvent, on the contrary, was labeled via crowdsourcing via
Amazon Mechanical Turk. It contains a total of 8,409 tweets in Spanish
and 7,303 in English, based on events related to different topics such
as entertainment, events, politics, global commemoration, and global
strikes. The labels that we use from this corpus correspond to Ekman's
basic emotions, complemented by `other'. DailyDialog, developed by
\citet{li2017dailydialog}, is a corpus consisting of 13,118 sentences
reflecting the daily communication style and covering various topics
related to daily life. The dialogues in the dataset cover totally ten
topics. It was annotated following Ekman's emotions by domain experts.
The ISEAR dataset was collected in the 90s by Klaus R. Scherer and
Harald Wallbott by asking people to report on their experience of
emotion-eliciting events. \cite{Scherer1997}. The dataset contains a
total of 7,665 sentences from 3,000 participant reports labeled with
single emotions. The last dataset that we use,
CrowdFlower\footnote{\url{https://data.world/crowdflower/sentiment-analysis-in-text}},
consists of 39,740 tweets labeled for 13 emotions. It is quite large,
but more noisy than some other corpora, given the annotation procedure
via crowdsourcing.

\subsection{Sentiment Analysis}

Sentiment analysis (SA) has emerged as one of the most well-known
areas in NLP due to its significant implications in social media
mining. Construed broadly, the task includes sentiment polarity
classification, identifying the sentiment target or topic, opinion
holder identification, and identifying the sentiment of one specific
aspect (e.g., a product, topic, or organization) in its context
sentence \cite{liu2010sentiment,abbasi2008sentiment}. Sentiment
analysis is a stricter sense, i.e.,~polarity classification, is often
modeled as a two-class (positive, negative) or three-class (positive,
negative, neutral) categorization task. For instance, the opinionated
expression ``The movie was terrible, I wasted my time watching it'' is
clearly negative. A negative sentiment can be an indicator of the
presence of offensive language, as previous studies have shown
\cite{elmadany2020leveraging,rodriguez2019automatic}. Sentiment
analysis and the identification of HOF share common discursive
properties. Considering the example shown in Section~\ref{sec:ea}, ``I
am sick and tired of this stupid situation'', in addition to
expressing anger, conveys a negative sentiment along with the presence
of expletive language targeted to a situation. Therefore, both
sentiment and emotion features can be used as useful information in
the NLP systems to benefit the task of HOF detection in social
media. Note that sentiment analysis is not a ``simplified'' version of
emotion analysis -- sentiment analysis is about the expression of an
opinion, while emotion analysis is about inferring an emotional
private state of a user. These tasks are related, but at least to some
degree complementary \cite{schuff-etal-2017-annotation}.

Unlike EA, as SA classification is one of the most studied tasks due
to its broader applications, a larger number of corpora annotated with
sentiments is available, particularly from Twitter. For instance, one
of the most well-known datasets is the Stanford Sentiment Treebank
\cite{socher-etal-2013-recursive}. It contains movie reviews in
English from Rotten Tomatoes. Another popular dataset was released in
SemEval 2016 for Task 4 is labeled with positive, negative, or neutral
sentiments and includes a mixture of entities (e.g., Gaddafi, Steve
Jobs), products (e.g., kindle, android phone), and events (e.g., Japan
earthquake, NHL playoffs) \cite{nakov-etal-2016-semeval}. The same
year, another dataset was released in SemEval for Task 6
\cite{mohammad-etal-2016-semeval}, the Twitter stance and sentiment
corpus which is composed of 4,870 English tweets labeled with positive
and negative sentiments.  For a more detailed overview, we refer the
reader to recent surveys on the topic
\cite{medhat2014sentiment,chakraborty2020survey,barnes-etal-2017-assessing}.

\subsection{Hate Speech and Offensive Language Detection}
Hate speech and offensive language (HOF) is a phenomenon that can be
observed with increasing frequency in social media in the last
years. As HOF became more widespread, attention from the NLP community
also increased substantially \cite{Tontodimamma2021}. Most research is
targeting Twitter as a platform, due to its popularity across various
user groups and its relatively liberal API terms and conditions for
researchers.  The methodological approaches vary between lexicon-based
approaches (among others, \cite{Gitari2015})
which are preferable due to its transparency in decision-making and
machine learning methods which typically show higher performance
\cite[i.a.]{Djuric2015,Badjatiya2017}. Definitions of hate speech for
the operationalization in automatic detection systems vary and do not
always strictly follow the ECRI definition that we introduced in
Section~\ref{sec:intro}. Sometimes, different categories such as hate
speech, profanity, offensive language, and abuse are collapsed into
one class because they are related \cite{Schmidt2017} and might
trigger similar responses (e.g., by authorities).

An important role in the HOF field has been played by a series of
shared tasks. One of the first of such events was GermEval
\cite{benikova2014germeval}, which was organized for the first time in
2014 and focuses on German. After initially focusing on information
extraction tasks, the identification of offensive language was
introduced in 2018 \cite{Wiegand2018}. Two subtasks were offered, one
on the classification into a non-offensive `other', and `profanity',
`insult', and `abuse', and one, in which the three latter classes are
collapsed to obtain a coarse-grained binary classification setup. This
setup was retained in the 2019 edition of the shared task
\cite{wiegand2019overview}.

Another well-known shared task event is OffensEval, which was held as
part of the International Workshop on Semantic Evaluation (SemEval) in
2019 and 2020
\cite{zampieri-etal-2020-semeval,zampieri-etal-2019-semeval}. As part
of the last OffensEval event, the OLID dataset was published, which
contains a total of 14,000 tweets in English. It was annotated using a
three-level hierarchical annotation scheme by crowdsourcing.

A third shared task series that took place in 2021 for the third time
is HASOC (Hate Speech and Offensive Content Identification in English
and Indo-Aryan Languages) \cite{mandl2019overview,Mandl2020}. In the
first edition of the HASOC, in 2019 \cite{mandl2019overview}, Hindi,
German and English datasets were created for the definition of HOF
based on Twitter and Facebook posts. HASOC 2020 introduced two tasks,
one on coarse-grained HOF vs.\ non-HOF language and one which
distinguishes hate, offensive language, and profane language for all
these languages. HASOC 2021 was extended by a subtask on code-mixed
language. This paper is a participation and system description in the
coarse-grained identification of HOF in English (Subtask 1A), in the
2021 edition of
HASOC\footnote{\url{https://hasocfire.github.io/hasoc/2021/}}.

Another direction of research relevant for this paper is formed by
studies that aim at obtaining a better understanding of the challenges
of HOF detection.  \citet{davidson2017automated} focused on the
separation between the classes of hate speech and offensive
language. They collected 33,458 tweets based on a crowdsourced lexicon
and found, based on bag-of-words maximum entropy classifiers, that
racist and homophobic tweets are more likely to be classified as hate
speech, but that sexist tweets are generally classified as offensive.
With a similar goal in mind, to understand which cases are
particularly challenging for HOF detection models,
\citet{rottger-etal-2021-hatecheck} introduced \textsc{HateCheck}, a
suite of functional tests, to enable more detailed insights of where
models might fail. They particularly analyzed distinct expressions of
hate, like derogatory hate speech, threatening language, slurs, and
profanity. Finally, \citet{Waseem2016} perform a corpus study to
understand which properties hate speech exhibits in contrast to
non-hateful language. This work is noteworthy because it focuses on
properties that are grounded in theories from social sciences rather
than being primarily data-driven.

Other research focused on the development of well-performing models
for HOF detection with adaptations to recent approaches to text
classification via transfer learning. \citet{mathur2018did}
investigated the usage of mixed language. They presented the
Multi-Input Multi-Channel Transfer Learning-based model (MIMCT) to
detect HS, offensiveness, and abusive language tweets from the
proposed Hinglish Offensive Tweet (HOT) dataset using transfer
learning coupled with multiple feature inputs. They stated that their
proposed MIMCT model outperforms basic supervised classification
models. \citet{wiedemann2018transfer}, participant in the GermEval
competition, used a different strategy for automatic offensive
language classification on German Twitter data. For this task, they
used a set of BiLSTM and CNN neural networks and include background
knowledge in the form of topics into the models.

We refer the interested reader to recent surveys on the topic of hate
speech and offensive language detection for a more comprehensive
overview \citep{Tontodimamma2021,MacAvaney2019}.

\subsection{Target Classification}
\label{sec:target}

According to the definition of hate speech, it must be targeted at a
particular individual or group, whether that target is mentioned
explicitly or not. Typical examples include black people, women, LGBT
individuals or people of a particular religion
\cite{lingiardi2020mapping}. The majority of current studies do not
aim at detecting targets that are mentioned in the text (in the sense
of information extraction), but aim at analyzing the properties of HOF
towards a particular group by sampling only posts aimed at that group
from social media. For example, \citet{kwok2013locate} focused on the
analysis of hate speech towards people of color, while
\citet{Grimminger2021} analyzed hate and offensive language by and
towards supporters of particular political parties.

Some studies aim at answering the research question how various target
groups are referred to. As an example, \citet{lemmens2021improving}
analyzed the language of hateful Dutch comments regarding classes of
metaphoric terms, including body parts, products, animals, or mental
conditions. Such a closed world approach, however, does not permit the
identification of targets that were not known at the development time
of the HOF detection system. This is to some degree addressed by
\citet{silva2016analyzing}, who developed a rule-based method to identify
target mentions that are then, similarly to
\citet{lemmens2021improving}, compared regarding the expressions that
are used.

\citet{elsherief2018hate} focused on the distinction between directed
hate, towards a particular individual as a representative of a group,
and generalized hate which mentions the group itself. Their study is
not focused on target classification, but on the analysis of which
groups and individuals are particularly in focus of hate speech,
including religious groups, genders, and ethnicities. To be able to do
that, however, they needed to automatically detect words in context of
HOF. They did that with the use of a mixed-effect topic model
\cite{Eisenstein2011}.

This label set is also used in the shared task OffensEval 2019
\cite{zampieri-etal-2019-semeval}, which is the only competition we
are aware of which included target classification as a subtask. The
OLID dataset of OffensEval 2019 has, next to HOF annotations, labels
which indicate whether the target is an individual, a group, some
other target, or if it omitted. We use this annotation in our study.

\section{Model}

Maybe the most pertinent question arising from our intuition above --
namely that HOF detection is related to the tasks of emotion,
sentiment and target classification -- is how this intuition can be
operationalized as a computational architecture. Generally speaking,
this is a \textit{transfer learning} problem, that is, a problem which
involves generalization of models across tasks and/or domains. There
are a number of strategies to address transfer learning problems; see
\citet{ruder2019neural} for a taxonomy. Structurally, our setup falls
into the inductive transfer learning category, where we consider
different tasks and have labeled data for each.  Procedurally, we
propose to learn the different tasks simultaneously, which amounts of
multi-task learning (MTL). In the MTL scenario, multiple tasks are
learned in parallel while using a shared representation
\cite{caruana1997multitask}. In comparison to learning multiple tasks
individually, this joint learning effectively increases the sample
size while training a model, which leads to improved performance by
increasing the generalization of the model \cite{zhang2017survey}.

\begin{figure}[t]
  \centering
  \includegraphics[width=0.95\linewidth]{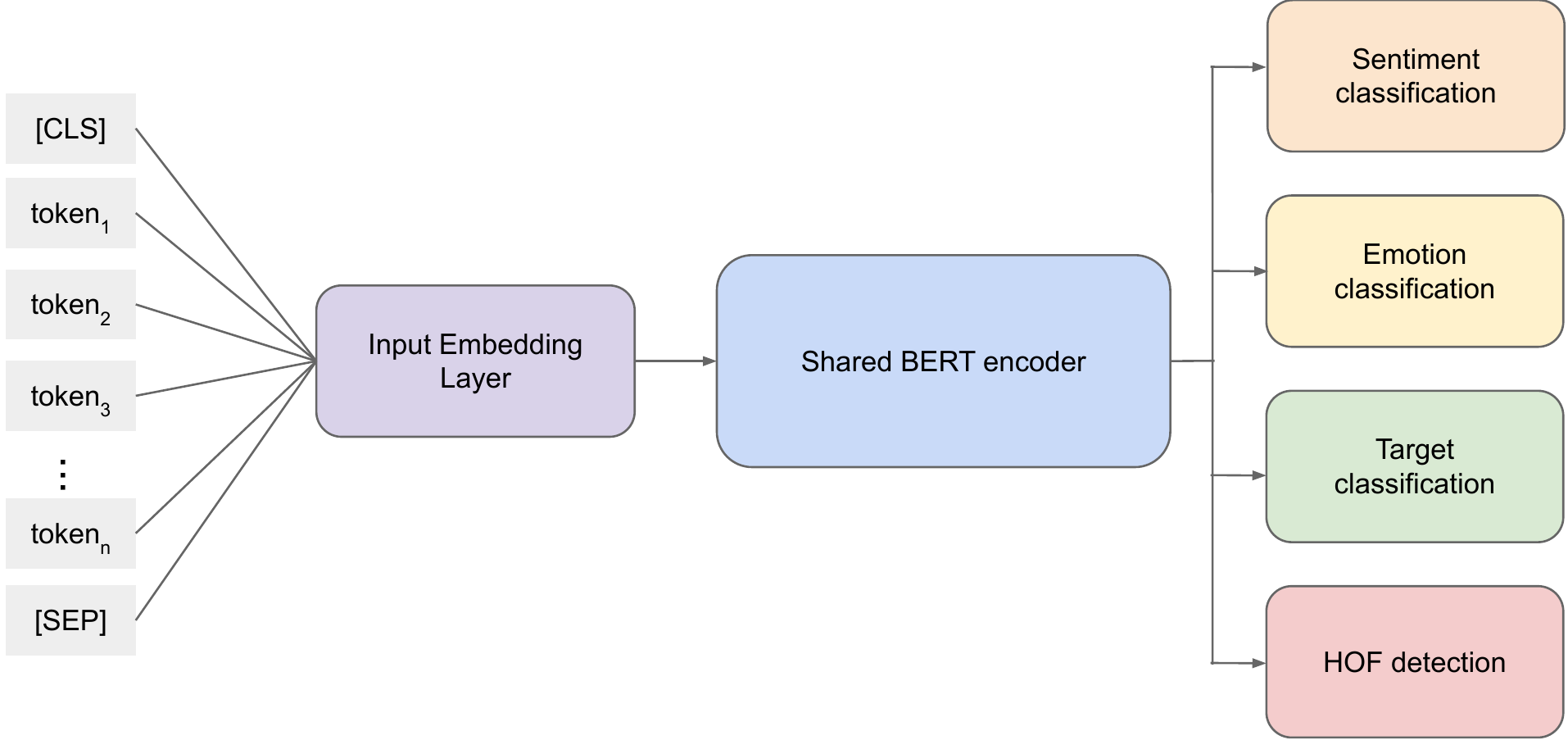}
  \caption{Proposed multi-task learning system to evaluate the impact
    of including emotion, sentiment, and target classification. The
    input representation is BERT-based tokenization and each task
    corresponds to one classification head. Information can flow from
    one task to another through the shared encoder that is updated
    during training via backpropagation.}
  \label{fig:multitask}
\end{figure}

The concrete MTL architecture that we use is shown in
Figure~\ref{fig:multitask}. We build on a standard contextualized
embedding setup where the input is represented by a transformer-based
encoder, BERT, pre-trained on a very large English corpus
\cite{Devlin2019}. We add four sequence classification heads to the
encoder, one for each task, and fine-tune the model on the four tasks
in question (binary/multiclass classification tasks). For the
sentiment classification task a tweet is categorized into positive and
negative categories; emotion classification classifies a tweet into
different emotion categories (\textit{anger}, \textit{disgust},
\textit{fear}, \textit{joy}, \textit{sadness}, \textit{surprise},
\textit{enthusiasm}, \textit{fun}, \textit{hate}, \textit{neutral},
\textit{love}, \textit{boredom}, \textit{relief},
\textit{none}). Different subsets of these categories are considered
in this task depending on the emotion corpus that is used to represent
the concept of an emotion. Target classification categorize the target
of the offense to an \textit{individual}, \textit{group}, to
\textit{others} and to be \textit{not mentioned}; and HOF detection
classifies a tweet into \textit{HOF} or \textit{non-HOF}. While
training, the objective function weights each task equally. At
prediction time, for each tweet in the HASOC dataset, four predictions
are assigned, one for each task.

\section{Experimental Setup}
\subsection{Experimental Procedure}

Our main research question is whether HOF detection can be improved by
joint training with sentiment, emotion and target. Even the adoption
of the architecture described in Section~3 leaves open a number of
design choices, which makes a model selection procedure necessary.

For the purpose of model selection, we decided to use the dataset
provided by the 2019 edition of the HASOC shared task, under the
assumption that the datasets are fundamentally similar (we also
experimented with the HASOC 2020 dataset, but the results indicated
that this dataset is sampled from a different distribution than the
2021 dataset). During the evaluation phase, we then used the best
model configurations we identified on HASOC 2019 to train a model on
the HASOC 2021 training data and produce predictions for the HASOC
2021 test set.

The two main remaining model selection decisions are (a), which
corpora to use to train the components?; (b), which components to
include? In the following, we first provide details on the corpora we
considered, addressing (a). We also describe the details of data
preprocessing, training regimen and hyperparameter handling. The
results are reported in Section 5 to address point (b).

\subsection{Corpora}
We carry out MTL experiments to predict HOF jointly with the concepts
of emotion, sentiment and HOF target. The datasets are listed in Table
\ref{tab:resources_dataset}.  To represent \textit{sentiment} in our
MTL experiments, we use the SemEval 2016 Task 6 dataset
\cite{mohammad-etal-2016-semeval} composed of 4,870 tweets in total. We
include the task of \textit{target classification} with the OLID
dataset \cite{zampieri-etal-2019-predicting}, which consists of 14,100
English Tweets. The concept of \textit{HOF} is modelled based on the
HASOC 2021 dataset, which provides three sub-tasks. We participate as
the team IMS-SINAI in sub-task 1A, which contains 5,214 English tweets
splits into 3,074 tweets in the training set, 769 in the development
set and 1,281 in the test
set.\footnote{\url{https://hasocfire.github.io/hasoc/2021/call_for_participation.html}}

For \textit{emotion detection}, we consider a set of six corpora in
the model selection experiment. These are the Crowdflower
data\footnote{\url{https://www.crowdflower.com/data/sentiment-analysis-emotion-text/}},
the TEC corpus \cite{mohammad-2012-emotional}, the Grounded Emotions
corpus \cite{liu2017grounded}, EmoEvent
\cite{plaza-del-arco-etal-2020-emoevent},
DailyDialogues\cite{li2017dailydialog}, and ISEAR. Among the available
emotion corpora, we chose those because they cover a range of
general topics and/or the genre of tweets.

\begin{table}[t]
\centering
\caption{Selection of resources for EA, SA and offensive
  target. The data sets that we use in our final experiments are
  marked with a star$^*$.}
\label{tab:resources_dataset}
  \begin{tabularx}{\linewidth}{llXll}
    \toprule
    \textbf{Category} & \textbf{Dataset} & \textbf{Annotation} & \textbf{Size} & \textbf{Source} \\ 
    \cmidrule(r){1-1}\cmidrule(rl){2-2}\cmidrule(lr){3-3}\cmidrule(lr){4-4}\cmidrule(l){5-5}
    \multirow{6}{*}{Emotion} & CrowdFlower$^*$ & Ekman's emo. & 39,740 & CrowdFlower (2016) \\
    & TEC & Ekman's emo & 21,051 & Mohammad (2012) \\
    & GroundedEmo. & sadness/joy & 2,585 & Liu et al. (2017) \\
    & EmoEvent & Ekman's emo/other & 7,303 & Plaza-del-Arco et al. (2020) \\
    & DailyDialogues & Ekman's emo & 13,118 & Li et al. (2017) \\
    & ISEAR & Ekman's emo/shame/guilt & 7,665 & Scherer (1994) \\
   Sentiment & SemEval 2016$^*$ &neg./pos./neutr. & 63,192 & Mohammad, Saif M. (2017) \\
   HOF & HASOC 2021$^*$ & Non/HOF & 5,124 & HASOC (2021) \\
    Target & OLID$^*$ & None/ind./group/other & 14,200 & OffensEval (2019) \\
    \bottomrule
  \end{tabularx}
\end{table}

\subsection{Data Preprocessing}
Tweets present numerous challenges in their tokenization, such as user
mentions, hashtags, emojis, misspellings, among others. To address
these challenges, we make use of the ekphrasis Python
library\footnote{\url{https://github.com/cbaziotis/ekphrasis}}
\cite{baziotis-pelekis-doulkeridis:2017:SemEval2}. Particularly, we
normalize all mentions of URLs, emails, users’ mentions, percentages,
monetary amounts, time and date expressions, and phone numbers. For
example, ``@user'' is replaced by the token ``<user>''. We further
normalize hashtags and split them into their constituent words. As an
example, ``\#CovidVaccine'' is replaced by ``Covid Vaccine''. Further,
we replace emojis by their aliases. For instance, the emoji
\raisebox{-1mm}{\includegraphics{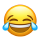}}\ is replaced by the token
``\textit{:face\_with\_tears\_joy:}'' using the emoji Python
library\footnote{\url{https://pypi.org/project/emoji/}}. Finally, we
replace multiple consecutive spaces by single spaces and replace line
breaks by a space.

\subsection{Training Regimen and Hyper-parameters} 

In the MTL stage, during each epoch, a mini-batch $b_t$ is selected
among all 4 tasks, and the model is updated according to the
task-specific objective for the task \textit{t}. This approximately
optimizes the sum of all multi-task objectives. As we are dealing with
sequence classification tasks, a standard cross-entropy loss function is used
as the objective.

\paragraph{Hyper-parameters.}
For hyper-parameter optimization, we split the HASOC 2021 into train
(80\,\%) and validation data (20\,\%).  Afterwards, in the evaluation
phase we use the complete training set of HASOC 2021 in order to take
advantage of having more labeled data to train our models.  For the
baseline BERT, we fine-tuned the model for four epochs, the learning
rate was set to $4\cdot10^{-4}$ and the batch size to 32. For
HASOC\_sentiment and HASOC\_emotion, we fine-tuned the model for three
epochs, the learning rate was set to $3\cdot10^{-5}$ and
$4\cdot10^{-5}$ respectively, and the batch size to 32. For
HASOC\_target, the epochs were set to four, the learning rate to
$4\cdot10^{-5}$ and the batch size to 16. For HASOC\_all, we
fine-tuned the model for two epochs, the learning rate was set to
$3\cdot10^{-4}$ and the batch size to 16. All the configurations used
AdamW as optimizer.

We run all experiments with the PyTorch high-performance deep learning
library \cite{paszke2019pytorch} on a compute node equipped with a
single Tesla-V100 GPU with 32 GB of memory.

\section{Results}
In this section, we present the results obtained by the systems we
developed as part of our participation in HASOC 2021 English subtask
1. We use the official competition metric macro averaged precision,
recall and \F-score as evaluation measures and further report
HOF-specific results, as we believe that, for real-world applications,
the detection of the concept HOF is more important than non-HOF.
The experiments are performed in two phases: the model selection
phase and the evaluation phase, which are explained in the following two
sections.

\begin{table*}[t]
\centering
\caption{MTL results for HOF detection on  HASOC 2019 test, varying
  the emotion dataset}
\label{tab:mtl_19hasoc_emotion}
\begin{tabular}{l ccc}
  \toprule
  &  \multicolumn{3}{c}{Macro Average} \\
  \cmidrule(r){2-4}
Emotion Dataset & P & R & \F  \\
\cmidrule(l){1-1} \cmidrule(l){2-2} \cmidrule(l){3-3} \cmidrule(l){4-4} 
TEC                  &   0.7583 &  0.7900 & 0.7707         \\
Grounded-Emotions    &   0.7744 &  0.7738 & 0.7741         \\
EmoEvent             &   0.7739 &  0.7807 & 0.7772        \\
DailyDialogs         &   0.7715 &  0.7865 & 0.7783       \\
ISEAR                &   0.7686 &  \textbf{0.7917} & 0.7785       \\
CrowdFlower          &   \textbf{0.7981} &  0.7778 &\textbf{0.7870}        \\
\bottomrule
\end{tabular}
\end{table*}

\begin{table*}[t]
\centering
\caption{MTL results for HOF detection on HASOC 2019 test set.}
\label{tab:mtl_19hasoc}
\setlength{\tabcolsep}{10pt}
\begin{tabular}{l l ccc ccc}
  \toprule
  && \multicolumn{3}{c}{Macro Average} & \multicolumn{3}{c}{Class HOF} \\
  \cmidrule(r){3-5}\cmidrule(r){6-8}
  Model &   & P & R & \F & P & R & \F \\
  \cmidrule(r){1-2}\cmidrule(r){3-5}\cmidrule(r){6-8}
Baseline & BERT & 0.775 & 0.779 & 0.777 & 0.66 & 0.674 & 0.667 \\
  \cmidrule(r){1-2}\cmidrule(r){3-5}\cmidrule(r){6-8}
MTL & HASOC\_sentiment   & 0.773          &  0.789          & 0.780          & 0.646 & 0.708 & 0.676 \\
    & HASOC\_emotion     & \textbf{0.798} &  0.778          & 0.787          & \textbf{0.712} & 0.642 & 0.675  \\
    & HASOC\_target      & 0.778          &  0.802          & 0.788          & 0.648 & \textbf{0.736} & 0.689   \\
    & HASOC\_all         & 0.791          &  \textbf{0.807} & \textbf{0.799} &  0.674 & 0.733 & \textbf{0.702}	  \\
\bottomrule
\end{tabular}
\end{table*}

\subsection{Model Selection (HASOC 2019)}

As described above, we perform model selection by training our
systems on the training set of HASOC 2019 and evaluating them on the
corresponding test set.  As our hypothesis is that the MTL system
trained on related tasks to HOF detection increased the generalization
of the model, we decided to use as a baseline the pre-trained language
model BERT fine-tuned on the HASOC 2019 corpora to compare the
results.

In order to decide which emotion corpora to use for the task of
emotion classification in the MTL setting, we test a number of emotion
datasets, obtaining the results shown in the Table
\ref{tab:mtl_19hasoc_emotion}.  These results are on the main task of
hate and offensive language detection, but vary the emotion dataset
used for MTL. As can be seen, the best performance is obtained by the
CrowdFlower dataset, with a substantial margin in terms of Macro-P
score. This is despite our impression that this dataset is comparably
noisy \cite{oberlander2018analysis}. We believe that what makes the
dataset suitable for HOF detection is that it contains a large number
of tweets labeled with a wide range of emotion tags, including
\textit{hate}. Therefore, we decided to use this emotion dataset in
the MTL setting for the final submission of HASOC 2021.

Table \ref{tab:mtl_19hasoc} shows the results of the MTL models
including the different auxiliary tasks on the HASOC 2019 test
data. The setting HASOC\_all refers to the MTL model trained on the
combination of all tasks (HOF detection, emotion classification,
polarity classification and offensive target classification). As can
be seen, the MTL models surpass the baseline BERT by at least 2
percentage points Macro-\F. In particular, the MTL model that obtains
the best performance is HASOC\_all, followed by HASOC\_target,
HASOC\_emotion and HASOC\_sentiment. The performance of HASOC\_all
increases by 2 points Macro-\F over the baseline, with Macro-Precision
increasing roughly 1.5 points and Macro-Recall roughly 2.5 points.

Table \ref{tab:mtl_19hasoc} further shows the results of the MTL
models on the HOF class in the HASOC 2019 test set. In all MTL systems
except HASOC\_emotion, the recall improved over the BERT baseline. The
highest improvement in terms of this measure is observed in the
HASOC\_target model, with an increase of 6.2 points. The precision
increases by 5.2 points in the HASOC\_emotion model. The best run
(HASOC\_all) outperforms the baseline BERT with a substantial margin
(0.702 to 0.667).

\begin{table*}[tb]
  \caption{\label{tab:knowledge} BERT vs. MTL predictions samples from
    HASOC 2019 test set, showing improved MTL performance. neg.:
    negative sentiment, pos.: positive sentiment, noemo: no emotion,
    ind.: individual target, None: not target detected}
  \centering
  \setlength{\tabcolsep}{4pt}
  \newcommand{\nhof}{$\overline{\text{HOF}}$}
  \begin{tabular}{lp{0.46\textwidth}@{\hskip 16pt}ccccccc}
    \toprule
    &&&& \multicolumn{4}{c}{MTL} \\
      \cmidrule(lr){5-8}
     ID & Tweet & Gold & BERT & {HOF} & {Sent.} & {Emot.} & {Targ.}\\
      \cmidrule(r){1-1}\cmidrule(lr){2-2}\cmidrule{3-8}
        107 & But Arlene and the extreme unionists do not want that, and they are the Jenga brick stopping the Tory roof collapsing & \nhof & HOF & \nhof & neg. & noemo & None  \\ [0.15cm]
        952 & I'ts his choice, you can't force him to get served by Muslims & \nhof & HOF & \nhof & neg. & noemo & None \\ [0.15cm]
        506 & Sad watching the UK making a total arse of itself & \nhof & HOF & \nhof & neg. & sadness & None \\ [0.15cm]
        4517 &	When you got average marks in exam... And ur Dad is like... dad.. She is so Beautiful. :- !	& \nhof & HOF & \nhof &	pos. & noemo & None \\ [0.15cm]
        254 & I don't think I have ever disliked anyone more than I dislike you. & HOF & \nhof & HOF & neg. & sadness & ind. \\ [0.15cm]
        684 & Yet you project the shortcomings of the muslim ruling class on to others, DEFLECTING, DIVERTING AND LYING TO THE MASSES!!! & HOF & \nhof & HOF & neg. & anger & None \\ [0.15cm]
        821 & Really, sounds like youre inviting open hostilities again. Are you sure your up to this job? Don't want to be rude but you're just not very bright and have a persistent habit of telling lies too. & HOF & \nhof & HOF & neg. & fear & ind. \\ [0.15cm]
        
    \bottomrule
  \end{tabular}
\end{table*}

\begin{table*}[t]
\centering
\caption{MTL results for HOF detection on HASOC 2021 dev set}
\label{tab:mtl_21hasoc_dev}
\begin{tabular}{llccc}
  \toprule
  &&  \multicolumn{3}{c}{Macro Average} \\
  \cmidrule(r){3-5}
  Model &   & P & R & \F \\
  \cmidrule(r){1-2}\cmidrule(l){3-5}
  Baseline & BERT & 0.801 & 0.796 & 0.798  \\
  \cmidrule(r){1-2}\cmidrule(l){3-5}
  MTL & HASOC\_sentiment            & 0.815	& 0.784 & 0.795   \\
& HASOC\_emotion              & 0.819 &	0.799 & 0.807   \\
& HASOC\_target               & 0.819 &	\textbf{0.802} & \textbf{0.809} \\
& HASOC\_all   &                \textbf{0.824} & 0.799 & \textbf{0.809}  \\
\bottomrule
\end{tabular}
\end{table*}

\paragraph{Model Analysis.}
As we aimed to improve HOF detection results by integrating the MTL
model with emotion, sentiment and target datasets, we decided to use
the pre-trained language model BERT in HASOC 2019 corpora as a basis
and compared the results of both BERT and MTL on HASOC\_all
models. The comparison of the two systems can be seen in Table
\ref{tab:knowledge}. Specifically, we show 7 examples, namely 4 false
positives and 3 false negatives performed by the baseline BERT
model. Regarding the false positives, the first two tweets (IDs 107
and 952) are predicted as HOF by the BERT model but MTL correctly
classified them as non-HOF, presumably because although the predicted
sentiment is negative, the model could neither recognize a negative
emotion nor a target to classify it as HOF. Tweet with ID 506 is also
correctly predicted by the MTL model as non-HOF, in this case,
although the emotion sadness is negative, we believe that it is not
strongly linked to HOF, moreover, the model does not recognize a
specific target directed at HOF. The last false positive (tweet ID
4517) expresses a positive sentiment and the model is able to
recognize it, thus we suppose that the MTL benefits from this
affective knowledge to classify the tweet as non-HOF. Regarding the
false negatives, the tweet with ID 254 has been classified by the MTL
system as negative sentiment, negative emotion (sadness) and is
directed to a person, therefore as these aspects are closely linked to
the presence of HOF, we assume that the MTL take advantage of these
aspects to correctly classify the tweet. The next sample, a tweet with
ID 684, expresses a negative opinion and an anger emotion, correctly
predicted by the MTL, this emotion is one of the emotions most
inextricably related to HOF, and together with the negative sentiment
could give a clue to the system to correctly classify the tweet as
HOF, although the target is not identified. Finally, instance
821 expresses a negative sentiment towards a person, correctly
identified by the MTL model. The model predicts fear for this instance
-- which we would consider a wrong classification. However, even from
this classification (fear instead of anger), the MTL model benefits
and makes the correct prediction, which was not possible in the plain
BERT model..

These examples indicate that our MTL system predicts the class HOF
more accurately than BERT and is particularly improved in cases that
have been missed by the plain model (which is also reflected by the
increased recall on the HASOC 2019 data).

\subsection{Model Evaluation (HASOC 2021)}

For evaluation, we use the dataset provided by the
organizers of the HASOC 2021 English subtask 1A. First, we want to verify
that the MTL models surpass the baseline BERT also in the evaluation
setting. We train all models on the HASOC 2021 training set and test
them on the dev set of HASOC 2021. The results obtained are shown in
Table \ref{tab:mtl_21hasoc_dev}.  As can be seen, most of the MTL
systems except HASOC\_sentiment outperform the baseline, which
validates our decision to select these models for the final evaluation
of HASOC 2021. HASOC\_sentiment does improve over the baseline in
Macro-Precision, but shows a drop in Macro-Recall. One reason might be
that the sentiment data that we use is in some relevant characteristic
more similar to the data from 2019 than to the data in the 2021
edition of the shared task.

Table \ref{tab:mtl_21hasoc_test} finally shows the five models that we
submitted to the HASOC 2021 Shared Task as team IMS-SINAI, both with
the official macro-average evaluation and the class-specific values
(which were reported during the submission period by the submission
system). We observe that BERT achieves a Macro-\F score of 0.790. The
multi-task learning models are, in contrast to the HASOC 2019 results,
mostly improved in terms of precision, and less consistently in terms
of recall. Considering the target classification and emotion
classification in multi-task learning models does not show any
improvements, however, the sentiment classification does. These
results for the separate concepts are contradicting the results on the
2019 data, which is an indicator that either the evaluation or
annotation procedures or the data has changed in some relevant
property: In the 2019 data, sentiment+HOF is not better than HOF, but
emotion+HOF and target+HOF are. In the 2021 data, it is vice versa.
However, when combining all concepts of sentiment, emotion, target,
and HOF in one model (HASOC\_all), we see an improvement that goes
above the contribution by the sentiment model alone. Therefore we
conclude that the concepts indeed are all helpful for the
identification of hate speech and offensive language.

\begin{table*}[t]
\centering
\caption{MTL results for HOF detection on HASOC 2021 test set
  (IMS-SINAI Team submissions). The official metric is the macro
  average score.}
\label{tab:mtl_21hasoc_test}
\setlength{\tabcolsep}{10pt}
\begin{tabular}{l l ccc ccc}
  \toprule
  && \multicolumn{3}{c}{Macro Average} & \multicolumn{3}{c}{Class HOF} \\
  \cmidrule(r){3-5}\cmidrule(r){6-8}
  Model &   & P & R & \F & P & R & \F \\
  \cmidrule(r){1-2}\cmidrule(r){3-5}\cmidrule(r){6-8}
  Baseline & BERT & 0.802 & 0.783 & 0.790 & \textbf{0.820} & 0.886 & 0.852 \\
  \cmidrule(r){1-2}\cmidrule(r){3-5}\cmidrule(r){6-8}
  MTL & HASOC\_sentiment  & 0.805 & 0.784 & 0.792 & 0.820 & 0.891 & 0.854  \\
& HASOC\_emotion  & 0.790 & 0.762 & 0.771 & 0.800 & 0.892 & 0.844  \\
& HASOC\_target & 0.800 & 0.776 & 0.785 & 0.813 & 0.892 & 0.851 \\
& HASOC\_all   & \textbf{0.819} & \textbf{0.784} & \textbf{0.795} & 0.812 & \textbf{0.917} & \textbf{0.862} \\
\bottomrule
\end{tabular}
\end{table*}

In addition, we report the results for the class HOF in the same
table, without averaging them with the class non-HOF. We find this
result particularly important, as the practical task of detecting hate
speech is more relevant than detecting non-hate speech.  The precision
values are lower than the recall values, in comparison to the average
results. The recall is particularly increased in the case of the best
model configuration (HASOC\_all) with 0.917 in comparison to 0.866 to
the plain BERT approach.  It is noteworthy that all multi-task models
increase the recall at the cost of precision for the class HOF. This
is both important for practical applications to detect hate speech in
the world and from a dataset perspective, as most resources have a
substantially lower label count of HOF than for other instances.

\section{Conclusion}
Most of the research conducted on the detection of hate speech and
offensive language (HOF) has focused on training automatic systems
specifically for this task, without considering other phenomena that
are arguably correlated with HOF and could therefore be beneficial to
recognize this type of phenomenon.

Our study builds on the assumption that the discourse of HOF involves
other affective components (notably emotion and sentiment), and is, by
definition, targeted to a person or group. Therefore, in this paper,
as part of our participation as IMS-SINAI team in the HASOC FIRE 2021
English Subtask1A, we explored if training a model concurrently for
all of these tasks (sentiment, emotion and target classification) via
multi-task learning is useful for the purpose of HOF detection. We
have used corpora labeled for each of the tasks, we have studied how
to combine these aspects in our model, and also we have explored which
combination of these concepts could be the most successful. Our
experiments show the utility of our enrichment method. In particular,
we find that the model that achieves the best performance in the final
evaluation considers the concepts of emotion, sentiment, and target
together. This improvement is even more clear in the HASOC 2019 data.
In an analysis of results, we have found that the model is good at
improving in false positives errors performed by BERT.  A plausible
mechanism here is that positive sentiments and positive emotions are
opposite to the general spirit of hate speech and offensive language
so that the presence of these indicators permit the model to predict
the absence of HOF more accurately.

This is in line with other previous results on multi-task learning
amongst multiple related tasks in the field of affective language. As
an example, \citet{akhtar-etal-2019-multi} has shown that both tasks
of sentiment and emotion benefit from each other. Similarly,
\citet{chauhan-etal-2020-sentiment} showed an improvement in sarcasm
detection when emotion and sentiment are additionally
considered. Particularly the latter study is an interesting result
that is in line with our work, because the sharp and sometimes
offending property of sarcasm is shared with hate speech and offensive
language. Further, \citet{rajamanickam-etal-2020-joint} has already
shown that abusive language and emotion prediction benefit from each
other in a multi-task learning setup. This also is in line with our
result, given that HOF is an umbrella concept that also subsumes
abusive language.

A clear downside of our model is its high resource requirement: it
needs annotated corpora for all the phenomena involved, and as our
model selection experiments showed, the quality of these resources is
very important. While resources that meet these needs are available
for English, for the vast majority of languages no comparable
resources exist. At the same time, the availability of multilingually
trained embeddings makes it possible to extend the transfer setup that
we adopted to a multilingual dimension, and train a model jointly on
resources from different languages. This perspective fell beyond the
scope of our study, but represents a clear avenue for future research,
and one that looks promising given the outcome of our
experiments. Other plausible extensions include the inclusion of
further affective phenomena that are arguably correlated to hate
speech, including stylistic ones such as sarcasm/irony
\cite{reyes2012humor} or author-based ones such as the "big five"
personality traits \cite{flek-2020-returning}; or a more detailed
modeling of the hate speech target beyond the coarse-grained
classification we used here, tying in, for example, with emotion role
labeling \cite{mohammad-etal-2014-semantic}.

Another aspect to study in more detail is based on the observation of
substantial differences between the results on the HASOC 2019 and the
HASOC 2021 data. Apparently, the improvements of the MTL model are
more clear on the 2019 data. This variance in results is an
opportunity to study the aspects that influence the performance
improvements when considering related concepts.

\section*{Acknowledgement}

This work has been partially supported by a grant from European Regional Development Fund (FEDER), LIVING-LANG project [RTI2018-094653-B-C21], and Ministry of Science, Innovation and Universities (scholarship [FPI-PRE2019-089310]) from the Spanish Government.

\bibliography{lit}

\end{document}